\documentclass[twocolumn,10pt]{article}

\usepackage{graphicx} 
\usepackage{booktabs} 
\usepackage{multirow}
\usepackage{rotating}
\usepackage{blindtext}
\usepackage{amsmath}
\usepackage{xcolor}
\usepackage{color}
\usepackage[linesnumbered,ruled]{algorithm2e}

\usepackage{pgfplotstable}
\usepackage{filecontents}

\usepackage{amssymb}

\usepackage{amsthm}
\usepackage{url}

\pgfplotstableread[col sep=semicolon]{data.txt}\data

\newcommand{\errplot}{%
\begin{tikzpicture}[trim axis right]
\definecolor{color1}{RGB}{235,163,104}
\begin{axis}[y=-\baselineskip,
  scale only axis,
  width=3.5cm,
  enlarge y limits={abs=0.5},
  axis y line*=middle,
  ytick=\empty,
  axis x line*=bottom,
  xbar,
  bar width=1.5ex,
  visualization depends on=x \as \rawx,
  nodes near coords,
  every node near coord/.style={
  anchor=east,
  shift={(axis direction cs:-\rawx,0)}
  }
 ]
\addplot [draw=black, fill=color1]
 table [x=Importance,y expr=\coordindex]{\data};
\end{axis}
\end{tikzpicture}%
}

\newcommand{\eg}{{\em e.g.}}
\newcommand{\etal}{{\em et~al.}}
\newcommand{\ie}{{\em i.e.}}
\newcommand{\etc}{{\em etc.}}

\newcommand{\name}{GraphSAGE}
\newcommand{\V}{\mathcal{V}}
\newcommand{\E}{\mathcal{E}}
\newcommand{\G}{\mathcal{G}}
\newcommand{\mb}{\mathbf}

\title{A Graph Neural Network Approach for Product Relationship Prediction}

\date{Paper accepted in ASME IDETC 2021}

\usepackage{authblk}

\author[1]{Faez Ahmed}
\author[2]{Yaxin Cui} 
\author[3]{Yan Fu}
\author[2]{Wei Chen}

\affil[1]{Dept. of Mechanical Engineering, Massachusetts Institute of Technology, Cambridge, MA}
\affil[2]{Dept. of Mechanical Engineering, Northwestern University, Evanston, IL}
\affil[3]{Insight and Analytics, Ford Motor Company, Dearborn, MI}

\begin{document}

\maketitle    

\begin{abstract}
{\it 
Graph Neural Networks have revolutionized many machine learning tasks in recent years, ranging from drug discovery, recommendation systems, image classification, social network analysis to natural language understanding. This paper shows their efficacy in modeling relationships between products and making predictions for unseen product networks. By representing products as nodes and their relationships as edges of a graph, we show how an inductive graph neural network approach, named GraphSAGE, can efficiently learn continuous representations for nodes and edges. These representations also capture product feature information such as price, brand, or engineering attributes. They are combined with a classification model for predicting the existence of the relationship between products. Using a case study of the Chinese car market, we find that our method yields double the prediction performance compared to an Exponential Random Graph Model-based method for predicting the co-consideration relationship between cars. While a vanilla GraphSAGE requires a partial network to make predictions, we introduce an `adjacency prediction model' to circumvent this limitation. This enables us to predict product relationships when no neighborhood information is known. Finally, we demonstrate how a permutation-based interpretability analysis can provide insights on how design attributes impact the predictions of relationships between products. This work provides a systematic method to predict the relationships between products in many different markets.
}
\end{abstract}

\section{Introduction}

Complex engineering systems contain multiple types of stakeholders and many individual entities, which exhibit complex interactions and interconnections. An example of a complex engineering system is the car market, where there are many interactions between stakeholders. The success of a new car depends not only on its engineering performance but also on the car's competitiveness relative to similar cars and factors such as perceived market position. Customers from different geographies may prefer different types of cars, while a design intervention in the car market, either by introducing changes in existing cars or by launching a new design of car may encourage customers to change their driving behavior. To solve this complexity, it is necessary to consider the complex relationship between customers and products, such as the social network between customers and the competitive relationship between products.

Network analysis has emerged as a key method for statistical analysis of engineering systems in a wide variety of scientific, social, and engineering domains~\cite{braha2006complex,holling2001understanding,hoyle2010integrated,newman2003structure,simon1977organization,wasserman1994social}. A few studies have begun exploring the capability of statistical network models in modeling  complex customer-product relationships~\cite{wang_analyzing_2015,fu_modeling_2017,sha_network-based_2018}. The premise underlying the network-based approach is that, similar to other engineering systems exhibiting dynamic, uncertain, and emerging behaviors, the relationship between customers and products can be viewed as a complex socio-technical system and analyzed using social network theories and techniques. The structural and topological characteristics identified in customer–product networks can reveal emerging patterns of the customer–product relations while taking into account the heterogeneity among customers and products.

Exponential random graph models (ERGMs) have been employed in literature as a statistical inference framework to interpret complex customer-product relations. ERGMs were used to study customers’ consideration behaviors using a unidimensional network at the aggregated market level~\cite{sha_analyzing_2017} and multidimensional network at the disaggregated customer level~\cite{wang_modeling_2016}, respectively. In unidimensional models, the product competition was established based on the customers consideration behaviour. The estimated unidimensional model was used to forecast the impact of technological changes (\eg{} turbo engines) on market competition~\cite{wang_forecasting_2016}, which illustrated the benefits of using the network-based preference model for predicting the outcome of design decisions.

However, ERGMs  have a few limitations. First, they are typically appropriate for small- to medium-sized networks with a few attributes. For large datasets, the MCMC approach to estimate ERGM parameters does not converge~\cite{cui2020weighted}. This leads to an important limitation for product manufacturers, who now want to make the most of huge datasets but still want statistical models that can help them understand what is happening under the hood. In addition, previously published research shows that future market forecasts based on  ERGMs  are not sufficiently accurate at capturing the true network~\cite{sha2018network}. In this paper, we provide an alternative approach of modeling networks using neural networks, which does not face these issues.

Graph neural network (GNN) are increasingly gaining popularity, given their expressive power and explicit representation of graphical data. Hence, they have a wide range of applications in domains that can harness graph structures out of their data. They offer fundamental advantages over more traditional unstructured methods in supporting interpretability, causality, and inductive generalization. Learning graph representations and performing reasoning and prediction has achieved impressive progress in applications ranging from drug discovery~\cite{stokes2020deep}, image classification, natural language processing and social network analysis~\cite{wu2020comprehensive}. A few of the well-known applications of GNNs are Uber Eats~\cite{uber}, who used them to recommend food items and restaurants and Alibaba using them to model millions of nodes for product recommendation~\cite{wang2018billion}. These successes motivated us to use them for studying product relationships.

We demonstrate a GNN approach for predicting product relationships. In our approach, the products are viewed as nodes, and the relationship among them (product association, market competition) is viewed as links. Hence, the problem of predicting relationships between products is posed as a graph link (or edge) prediction problem. 
The new approach we develop in this study is based on GraphSAGE, a type of GNN method, which allows modeling of design attributes. GraphSAGE first represents a graph (network) structure in lower-dimension vectors and utilizes the vectors as the downstream classification input. Meanwhile, we develop a permutation-based method to examine the feature importance to assist design decisions. In summary, the contributions of this study are:

\begin{enumerate}
    \item Propose a GNN-based method for modeling a product relationship network and enabling a systematic way to predict the relationship links between unseen products for future years.
    \item Show that the link prediction performance of GNNs is better than existing network modeling methods.
    \item Demonstrate the scalability of the GNN method by modeling the effect of a large number of continuous and categorical attributes on link prediction.
    \item Uncover the importance of attributes to help make design decisions using permutation-based methods.

\end{enumerate}

\section{Related Work}
This paper applies GNNs to product relationship networks for link prediction and uncovers the importance of engineering design attributes using permutation-based analysis. In this work, we focus on the product co-consideration relation as a demonstration, but the method can be generalized to other product relationships, such as product association relationship.
Below, we discuss related work on product co-consideration networks, GNNs, and interpretable machine learning.

\subsection{Product Co-consideration Networks}

Co-consideration of products describes the situation where customers consider multiple products at the same time prior to making a purchase~\cite{wang2018predicting}. The consideration behavior involves the comparison and evaluation of product alternatives and is accordingly a key step in the customer's decision-making process~\cite{shocker1991consideration}. At the same time, product co-consideration also means a market competition relationship between products, which is crucial to the company's product positioning plans and market strategies. As a single product may be chosen by a customer considering two or more products, those products can increase their market share by understanding competition relationships and introducing interventions for them to be preferred over their competitors.
Therefore, the successful modeling of the product co-consideration relationship helps enterprises understand the embedded market competition and provides new opportunities for enterprises to formulate design solutions to meet customer needs.

In order to understand the underlying patterns of customer consideration behaviors, researchers have developed multiple methods and models of customer considerations. Some models of customer consideration set composition are based on the marginal benefits of considering an additional product~\cite{hauser1990evaluation, roberts1991development}. Other pioneering works have built models for investigating the impact of the consideration stage on the customer decision-making process~\cite{gaskin2007two,dieckmann2009compensatory}.
Also, in the Big Data era, many works use both online and offline customer activity data to infer the product co-consideration behavior~\cite{damangir2018uncovering}. In recent years, the network-based approach has emerged to understand the product competition by describing the product co-consideration relation based on customer cross-shopping data~\cite{wang2018predicting,sha2018network,cui2020weighted}. Depicted in a simple network graph, where nodes represent individual products and edges represent their co-consideration relation based on aggregated customer preference, network-based analysis views co-consideration relations in terms of network theories, and the links in the observed network are explained by the underlying social processes.

\begin{table*}[ht!]
\centering
\caption{Comparison of this work with prior studies on modeling car relationship using ERGM models}
\label{tab:ERGM and this work}       
\begin{tabular}{p{2cm}p{6.5cm}p{6.5cm}}
\toprule
\textbf{Topic} & \textbf{Past work using ERGM model} &  \textbf{This work using GNN model} \\
\midrule
Train nodes & 296 cars (common cars between 2013 and 2014) & 388 cars (all cars from 2013)\\
Test nodes & Tested on 296 cars from 2014 & Tested on 403 cars from 2014 and 422 cars from 2015\\
Unseen data & Predictions restricted to cars in the training data & Predictions for completely new cars too (107 unseen cars in 2014)\\
Attributes & 6 numerical design attributes & 29 design attributes, including categorical attributes\\
Interpretability & Coefficient-based &  Permutation-analysis based\\

\bottomrule

\end{tabular}

\end{table*}

Several works that investigate the product co-consideration network are based on a dataset of car purchases. Wang \etal~\cite{wang2016network} have applied the correspondence analysis methods and network regression methods to investigate the formation of car co-consideration relationship. Sha \etal~\cite{sha2018network} have applied ERGMs in understanding the underlying customer preference in  car co-consideration networks. However, the previous explorations are restricted to using the traditional network-based statistical methods, which leads to a low computation efficiency, low prediction accuracy for the future market competition as well as inability to model many design attributes. To overcome the limitations of the ERGMs, we have developed a new method to investigate the underlying effect of customers' consideration behavior by using GNN methods. 
Applied to the same dataset, a comparison of the ERGMs  and this work is summarized in Table~\ref{tab:ERGM and this work}.

\subsection{Graph Neural Networks} 

Network data can be naturally represented by a graph structure that consists of nodes and links. Recently, research on analyzing graphs with machine learning has grown rapidly. The graph-based machine learning tasks in networks include \textbf{node classification} (predict a type of given node), \textbf{link prediction} (predict whether two nodes are linked), \textbf{community detection} (identify densely linked clusters of nodes), \textbf{network similarity} (determine how similar two networks are), \textbf{anomaly detection} (find the outlier nodes), and \textbf{attribute prediction} (predict the features of a node)~\cite{zhou2018graph}.

In a graph, each node is naturally defined by its features and the neighborhood of connected nodes. Therefore, learning the representation of nodes in a graph, called \textbf{node embedding}, is an essential part of  downstream tasks such as classification and regression. Most node embedding models are based on spectral decomposition~\cite{kipf2016semi,atwood2015diffusion} or matrix factorization methods~\cite{cao2016deep,qiu2018network}. However, most embedding frameworks are inherently transductive and can only generate embeddings for a single fixed graph. These transductive approaches do not efficiently generalize to unseen nodes (e.g., in evolving graphs), and these approaches cannot learn to generalize across different graphs. In contrast, \textbf{GraphSAGE} is an inductive framework that leverages node attribute information to efficiently generate representations on previously unseen data.
GraphSAGE samples and aggregates features from a node's local neighborhood ~\cite{hamilton2017inductive}. By training a GraphSAGE model on an example graph, one can generate node embeddings for previously unseen nodes as long as they have the same attribute schema as the training data. 
It is especially useful for graphs that have rich node attribute information, which is often the case for product networks.

\subsection{Interpretable Machine Learning}
In addition to using machine learning models for prediction, there is growing attention on the capability to interpret what a model has learned. Interpretable machine learning would be an effective tool to explain or present the model results in terms understandable to humans~\cite{doshi2017towards,molnar2020interpretable}.   

As a traditional machine learning explanation method, feature importance indicates the statistical contribution of each feature to the underlying model~\cite{du2019techniques}. Among the techniques to unravel the feature importance, model-agnostic interpretation methods~\cite{ribeiro2016should} treat a model as a black-box and do not inspect internal model parameters, which has the advantage that the interpretation method can work with any machine learning model. A representative approach is the \textbf{permutation feature importance measurement}, which was introduced by Breiman~\cite{breiman2001random} for random forests. Based on this idea, Fisher~\etal{}~\cite{fisher2019all} developed a model-agnostic version of the feature importance and called it model reliance. The key idea is that the importance of a specific feature to the overall performance of a model can be determined by calculating how the model prediction accuracy deviates after permuting the values of that feature~\cite{altmann2010permutation}. The permutation-based feature importance method has been applied to bioinformatics~\cite{putin2016deep}, engineering~\cite{matin2018variable}, and political science~\cite{farinosi2018innovative} to provide a highly compressed, global insight into machine learning models. In our study, we use the permutation-based methods to examine important product attributes that impact the link prediction between cars.

\section{Methodology}

\begin{figure*}[t]
\centering
\includegraphics[width=\textwidth]{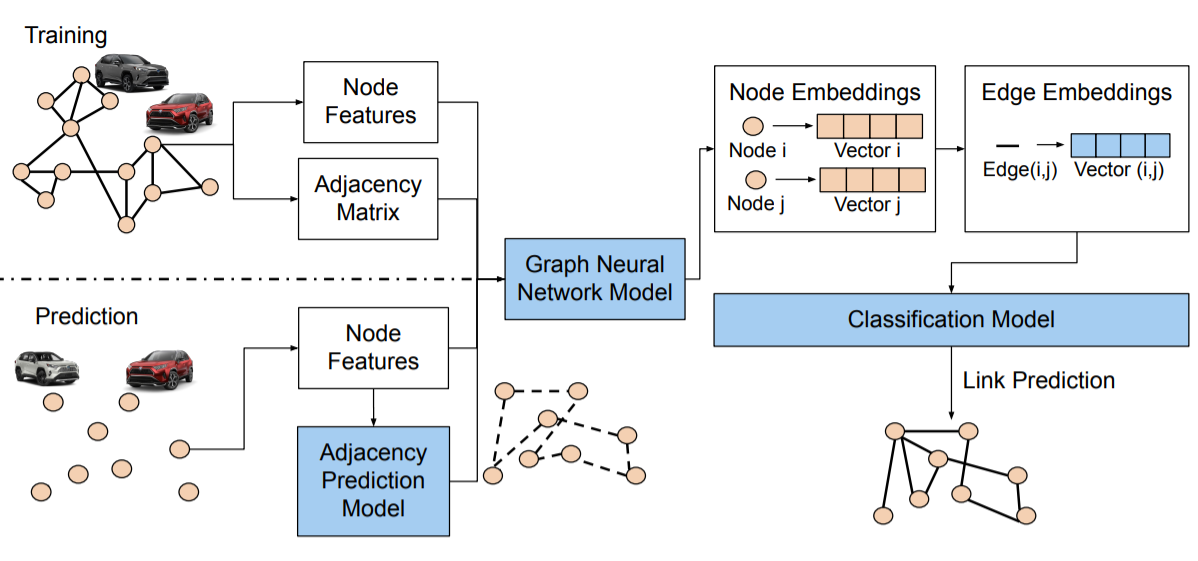}
\caption{The methodology of predicting the link existence in a car competition network using graph neural network model}
\label{fig:methodology}
\end{figure*}

We establish a product co-consideration network to model product competition behavior and use a GNN approach to predict future product competition. The methodology of the training and prediction process for the link existence is shown in Fig.\ref{fig:methodology}. 

Our methodology comprises five main components, which include representing products and their relationships as a graph, training the GNN to learn the graph structure, training classification models to make predictions, creating an adjacency prediction model to augment the GNN for unseen data, and, finally, interpreting the importance of design attributes. These components are described next:

\subsection{Network Construction}
\label{subsec:network construction}

Networks present a natural way to simultaneously model products as nodes and relationships between them as edges. Before purchasing a product, customers often consider multiple products and then select one or more products among them. When two products are simultaneously considered by the same customer in their decision-making process, we define this relationship as a co-consideration relationship. Assuming the customer only buys one product in the end, products that are co-considered are assumed to be in competition in this paper. Note that there are many different methods to measure competition between any two products, and the methods we describe next generalize to any measure of choice. Next, we discuss how a graph is created for co-considered products. Readers who already have a predefined network using some other method can skip this section.

We studied a unidimensional product network that can reveal product market competition by describing products' co-consideration relationship. Each product corresponds to a unique node. Each node is associated with a set of attributes such as price, fuel consumption, and engine power. 
The product co-consideration network is constructed using data from customers' consideration sets. The presence of a co-consideration binary link between two nodes (products) is determined by the number of customers who consider them together:

\begin{equation}
\label{link def}
 E_{i,j}=
\begin{cases}
    1, & n_{i,j} \geq cutoff\\
    0, & otherwise
\end{cases}   
\end{equation}

where $E_{i,j}$ refers to the edge connected by node $i$ and node $j$. $n_{i,j}$ is the number of customers who have considered products $i$ and $j$ together. $cutoff$ is a domain-dependent threshold, which defines the strength of the relationship considered in the analysis.
In other words, we define an undirected link between node $i$ and node $j$, if there exists at least one customer who considers both products $i$ and $j$ together. Based on Equation~\ref{link def}, the network adjacency matrix is symmetric and binary. This study uses a cut-off value equal to 1. 

\subsection{Inductive Representation Learning on Networks}
Finding a low-dimensional vector embedding of nodes and edges in graphs can enable many machine learning applications such as node classification, clustering, and link prediction. This section  describes a method to learn such embeddings, named GraphSAGE. GraphSAGE is a representation learning technique for dynamic graphs. It can predict the embedding of a new node, without needing a re-training procedure. To do this, GraphSAGE uses inductive learning. It learns aggregator functions that can induce new node embedding, based on the features and neighborhood of the node.

\begin{figure*}[t]
\centering
\includegraphics[width=\textwidth]{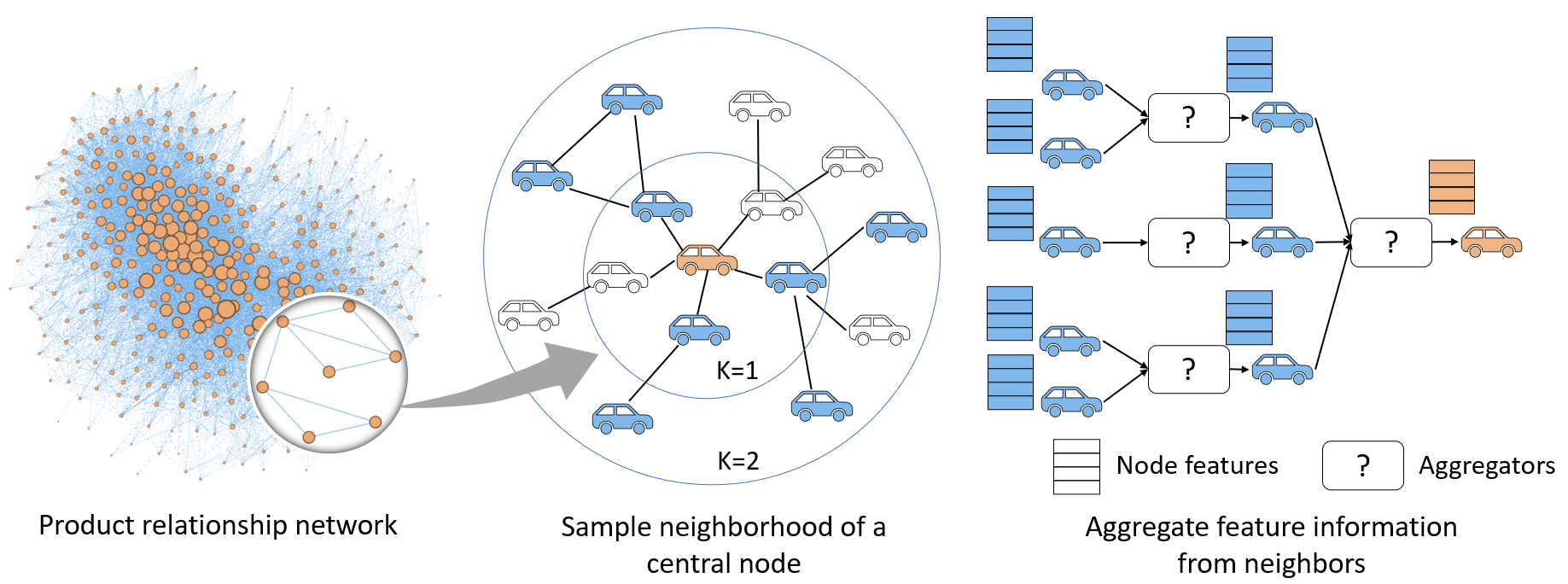}
\caption{Illustration of sampling and aggregation in GraphSAGE method. A sample of neighboring nodes contributes to the embedding of the central node.}
\label{fig:graghsage}
\end{figure*}

As illustrated in Fig.~\ref{fig:graghsage}, GraphSAGE learns node embeddings for attributed graphs (where nodes have features or attributes) through aggregating neighboring node attributes. The aggregation parameters are learned by encouraging node pairs co-occurring in short random walks to have similar representations. 
Many GNN models learn functions that generate the embeddings for a node, which sample and aggregate feature and topological information from the node's neighborhood. However, the benefit of training a GraphSAGE model, in contrast to other GNN methods, is its inductive behavior, which is necessary for engineering applications.
Most other GNN methods are transductive, which means they can only generate embeddings for a single fixed graph. If a completely new product comes up in a dynamically evolving graph, these transductive approaches cannot generalize to such unseen nodes. In contrast, GraphSAGE is an inductive method that leverages the attribute information of a new node to efficiently generate representations on previously unseen data. The detailed algorithm of GraphSAGE is shown in Algorithm ~\ref{alg:basic}. Interested readers are encouraged to read~\cite{hamilton2017inductive} for details of the algorithm.

\begin{algorithm}
\caption{\name\ embedding generation (i.e., forward propagation) algorithm from~\cite{hamilton2017inductive}}
\label{alg:basic}
	\SetKwInOut{Input}{Input}\SetKwInOut{Output}{Output}
    \Input{~Graph $\G(\V,\E)$; input features $\{\mb{x}_v, \forall v\in \V\}$; depth $K$; weight matrices $\mb{W}^{k}, \forall k \in \{1,...,K\}$; non-linearity $\sigma$; differentiable aggregator functions $\textsc{aggregate}_k, \forall k \in \{1,...,K\}$; neighborhood function $\mathcal{N} : v \rightarrow 2^{\V}$}
    \Output{~Vector representations $\mb{z}_v$ for all $v \in \V$}
    \BlankLine
    $\mb{h}^0_v \leftarrow \mb{x}_v, \forall v \in \V$ \;
    \For{$k=1...K$}{
    	  \For{$v \in \V$}{
    	  $\mb{h}^{k}_{\mathcal{N}(v)} \leftarrow \textsc{aggregate}_k(\{\mb{h}_u^{k-1}, \forall u \in \mathcal{N}(v)\})$\;
    	  		$\mb{h}^k_v \leftarrow \sigma\left(\mb{W}^{k}\cdot\textsc{concat}(\mb{h}_v^{k-1}, \mb{h}^{k}_{\mathcal{N}(v)})\right)$
    	  }
    	  $\mb{h}^{k}_v\leftarrow \mb{h}^{k}_v/ \|\mb{h}^{k}_v\|_2, \forall v \in \V$
    	}
     $\mb{z}_v\leftarrow \mb{h}^{K}_v, \forall v \in \V$ 
\end{algorithm}

To train a GraphSAGE model, the inputs are the product attributes (\ie~node features) and the network structure (\ie~adjacency matrix) of the product co-consideration network.
Then for each node, the GNN models are able to encode nodes into lower-dimensional space in the node embedding stage. For example, as illustrated in Fig.\ref{fig:methodology}, nodes $i$ and $j$ can be represented by vectors $i$ and $j$, which carry the information of node $i$'s and $j$'s features and local neighborhoods, respectively.

\paragraph{Edge embeddings} Using a GNN-trained embedding for nodes, one can also learn the representation for all possible links (edges) in the network. This is done by aggregating every possible pair of node embeddings. We use the dot product of vectors $i$ and $j$ to find the edge embeddings. Note that other symmetric operations such as addition can also be used to aggregate two node embeddings to give an edge embedding. In our experiments, we did not find significant differences among aggregation methods on the final prediction performance. After training, edges between similar nodes are expected to be closer to each other in the edge embedding space.

Once we learn the edge embeddings, they can be used as an input to any machine learning model, which can be trained to predict whether an edge exists or not, which is discussed next.

\subsection{Classification Model for Link Prediction}
The link prediction problem can be viewed as a binary classification problem, where the goal is to predict whether a link candidate exists in the network (Class 1 or a positive edge) or does not exist (Class 0 or a negative edge).
During the GNN model training, we can also train a downstream classification model to predict link existence, given the edge embedding as an input. 

For each pair of nodes, the classification model takes the edge embeddings as input and whether the link exists or not as labels. Any classification model, such as logistic regression, k-nearest neighbors, and naive Bayes classifiers, can be integrated with the GNN model to predict the link existence. We used a multilayer perceptron (MLP) model for this work. Note that in the training process, the GNN model and the classification model are trained simultaneously for the supervised learning task. To avoid imbalanced training of the classification model for networks with very few edges, the two classes are balanced by sub-sampling the negative edges (an edge which does not exist in the training data). 

\subsection{Validation Networks}
After the training was completed, we tested the performance of the model in predicting links for an unseen network. The model can be tested on two different types of networks. In one case, the initial network was divided into two parts by randomly sampling edges. The GNN model was tested to predict links for the held-out links. In the second case, we trained the model on one network and tested it on another completely unseen network. However, this presents new challenges, which are discussed next.

\subsection{Adjacency Prediction Model}

While GNN-based link prediction methods are typically used to find missing links from a graph, they cannot be directly applied to a completely unknown network. However, in engineering design applications, it is possible  to train a model on products in Year 1 and make predictions about Year 2, which may have new products and evolved versions of previous products. Applications may require that predictions for links between products are made where training and testing networks belong to different domains, time periods, or locations.
This presents a circularity problem, as a typical GNN, including GraphSAGE, needs atleast a partial adjacency matrix as an input, to predict the complete adjacency matrix.

We overcame this issue by developing a method to predict an approximate adjacency matrix using a separate machine learning model, which is referred to as the adjacency prediction model in Fig.\ref{fig:methodology}. The predicted adjacency is used to identify a few neighbors of each node, which are used in the GNN as a partial adjacency matrix. 

There are several ways of predicting the adjacency matrix, given the node attributes. A na\"ive way would be to find all the nodes in the new graph, which also appeared in the training dataset, and copy their adjacency information. However, such a model performs poorly, as all the new nodes have no neighbors, therefore, the GNN cannot make accurate predictions about them.
 
Instead, we used a similarity-based K-nearest neighbor method in the adjacency prediction model. The similarities among product nodes are measured by the cosine distance of all car features. Using these similarities for each node, K most similar nodes from the graph are selected as neighbors. This gives us the approximate adjacency matrix, where each node is connected to its K-nearest neighbors. The benefit of this approach is that all nodes in the co-consideration network are connected to some other nodes. While the choice of K is subject to the modeler, we seek an appropriate number to keep the density of the network comparable with a typical co-consideration product network in the training network.

Note that other machine learning methods can also be used to output an approximate adjacency matrix. For instance, one can train a classification model with the average car attributes as input and a binary output corresponding to link existence. Our preliminary analysis showed that classification models (\eg{} logistic regression) did not perform as well as the nearest-neighbor approach. This may be attributed to classification models not finding sufficient neighbors for all nodes. Our method overcame this limitation by assigning the same number of neighbors to all nodes, which also yields good empirical results.

\subsection{Metrics for Link Prediction}

With the trained GNN model and classification model, we predicted the co-consideration network in the subsequent years based on the new node features and the approximate adjacency prediction model. The link prediction can be regarded as a binary classification model, which predicts the probability of the target link existence to be Yes or No. 
To evaluate the performance of the classification model, we analyzed the confusion matrix (which describes the performance of a classifier) and the receiver operating characteristic (ROC) curve, which plots the true positive rate and false positive rate). To compare different models, we used the area under the curve (AUC) metric, which measures the area underneath the ROC curve, and provides an aggregated measure of the performance across all possible classification thresholds. The AUC ranges in value from 0 to 1, a higher AUC value indicates a better classification model.

\subsection{Permutation-based Feature Importance}

In the engineering design domain, besides forecasting the future market competition, it is important to understand the dominant features in product competition. Therefore, we investigated the importance of different design attributes in the GNN method.

Feature importance is the increase in model error when the feature's information is destroyed.
Permutation feature importance measures the increase in the prediction error of the model after we permuted the feature's values, which breaks the relationship between the feature and the true outcome. We measured the importance of a feature by calculating the increase in the model's prediction error after permuting the feature. A feature is ``important'' if shuffling its values increases the model error, because in this case the model relied on the feature for the prediction. A feature is ``unimportant'' if shuffling its values leaves the model error unchanged, because in this case the model ignored the feature for the prediction.

Feature importance based on the training data tells us which features are important for the model in the sense that it depends on them for making predictions.
Permutation feature importance does not require retraining the model. Some other methods suggest deleting a feature, retraining the model, and then comparing the model error. Since the retraining of a machine learning model can take a long time, "only" permuting a feature can save a lot of time.
Permutation feature importance is linked to the error of the model.

Outline of the permutation importance algorithm:

\begin{enumerate}
    \item Inputs: fitted predictive model $m$, tabular dataset (training or validation) $D$.
    \item Compute the reference score $s$ of the model $m$ on data $D$ using all features
    \item For each feature $j$ (column of $D$):
    \begin{enumerate}
        \item For each repetition $k$ in $1,...,K$
        \begin{enumerate}
            \item Randomly shuffle column $j$  of dataset $D$  to generate a corrupted version of the data named $\widetilde{D}_{k,j}$
        \item Compute the score $s_{k,j}$  of model on corrupted data $\widetilde{D}_{k,j}$
        \end{enumerate}

        \item Compute importance $i_j$ for feature $f_j$ defined as: $1-\frac{s_{k,j}}{s}$
    \end{enumerate}
\end{enumerate}

\section{Results and Discussion}

In this section, we demonstrate the use of the GNN approach to study the Chinese car market. We use car survey data provided by the Ford Motor Company as a test example. By training a network model, we can predict the future market competition even though car attributes are changing and new products are introduced. Meanwhile, the feature importance in the car competition network is examined for the training network, which can be reported back to designers to make strategic design changes. 

\subsection{Data Description}
Our dataset contains customer survey data from 2012 to 2016 in the China market. In the survey, there were more than 40,000 respondents each year who specified which cars they purchased and which cars they considered before making their final car purchase decision. 
Each customer indicated at least one and up to three cars which they considered. The dataset resulting from the survey also contains many attributes for each car (\eg{} price, power, brand origin, and fuel consumption) and many attributes for each customer (\eg{} gender, age). 

\subsection{Link Prediction for Car Co-Consideration Network}
In this example, we used our method to build a model that predicts co-consideration links in the car dataset aforementioned. The problem is treated as a supervised link prediction problem on a homogeneous network with nodes representing cars (with attributes such as engine size and categorical body type) and links corresponding to car-car co-consideration relationship.

\paragraph{Network construction}
To study car co-consideration, we started by creating a car co-consideration network based on customers' survey responses in the 2013 survey data. The network consists of 388 unique car models as network nodes. The link between a pair of nodes (denoting cars) is all allocated based on the presence of the car co-consideration by at least one customer.

\paragraph{The input car attributes}
As demonstrated in the Methodology section, the car attributes and co-consideration network adjacency matrix serve as the input of the GNN and classification models, and the link existences are labels to judge the training performance. Our experiment studied 29 car attributes, which were manually chosen. The list of attributes contain all the effective engineering attributes (\eg~fuel consumption, engine size) and car types (\eg~body type, market segmentation) available in the survey dataset. The attributes are listed in Table~\ref{tab:featureimportance}. Note that the attributes are both continuous and categorical. The categorical variables are transformed via a one-hot encoder which converts categorical variables into vectors (after one-hot encoding 29 features lead to 210 features), and the continuous variables are normalized to vary between 0 and 1. 

\paragraph{Experimental settings}
In the training process, we built a model with the following architecture. First, we built a two-layer GraphSAGE model that takes labeled node pairs corresponding to possible co-consideration links, and outputs a pair of node embeddings for the two cars of the pair. These embeddings were then fed into a link classification layer, which first applied a binary operator to those node embeddings (dot product) to construct the embedding of the potential link. The thus-obtained link embeddings are passed through the dense link classification layer to obtain link predictions - the probability for these candidate links to actually exist in the network. The entire model was trained end-to-end by minimizing the loss function of choice (e.g., binary cross-entropy between predicted link probabilities and true link labels, with true/false links having labels 1/0) using stochastic gradient descent (SGD) updates of the model parameters, with minibatches of training links fed into the model.

We specified the minibatch size (number of node pairs per minibatch) and the number of epochs for training the model to 20 and 100. As for the number of sampled neighbors, we set the sizes of 1-hop and 2-hop neighbor samples for GraphSAGE to be 20 and 10. For the GraphSAGE part of the model, we selected hidden layer sizes of 20 for both GraphSAGE layers and a bias term, and a dropout rate of 0.3. We stacked the GraphSAGE and prediction layers and defined the binary cross-entropy as the loss function. Parameters were chosen based on initial analysis on a validation set. Our code will be made public on Github for other researchers to replicate our results.

\paragraph{Predicting missing links in the 2013 network}
We split our input graph into a training graph and a test graph. We used the training graph for training the model (a binary classifier that, given two nodes, predicts whether a link between these two nodes should exist or not) and the test graph for evaluating the model’s performance on hold-out data. Each of these graphs will have the same number of nodes as the input graph, but the number of links will differ (be reduced) as some of the links will be removed during each split and used as the positive samples for training/testing the link prediction classifier.

\begin{table*}[t]
    \centering
    \caption{Confusion matrix in predicting 2013 with 29 features. Average F1-score for 2013 is 0.74. AUC for 2013 train is 0.84 and test is 0.84. True Negative Rate (TNR) and True Positive Rate (TPR) are shown in brackets.}
    
    \begin{tabular}{@{}cccccc@{}}
    \multicolumn{2}{c}{}   &  \multicolumn{2}{c}{2013 training prediction} &
    \multicolumn{2}{c}{2013 test prediction on held-out links}\\
    \noalign{\smallskip}\hline\noalign{\smallskip}
         \multirow{6}{*}{\begin{sideways} Actual Class \end{sideways}} &  &  0 &  1 & 0 & 1 \\
\hline\noalign{\smallskip}
         \rule{0pt}{23pt} & 0 & 5390 (TNR 53.90\%) & 4610 (FPR 46.10\%)  & 609 (TNR 54.82\%) & 502 (FPR 45.18\%)\\
         \rule{0pt}{23pt} & 1 & 592 (FNR 5.92\%) & 9408 (TPR 94.08\%) & 75 (FNR 6.75\%) & 1036 (TPR 93.25\%)\\
\noalign{\smallskip}\hline\noalign{\smallskip}
    \end{tabular}
    \label{tab:confusion_matrix_2013}
\end{table*}

The prediction performance along with the training performance is first measured by a confusion matrix in Table~\ref{tab:confusion_matrix_2013}. The right-hand part of Table~\ref{tab:confusion_matrix_2013} shows the confusion matrix of 2013 test prediction on held-out links. It includes 4 different combinations of predicted and actual classes. The 609 in the top-left cell is the true negative (predict negative and it's true), the 502 in the top right is the false positive (predict positive and it's false). The associated percentages indicate that for all pairs of nodes without link existence (actual class = 0), 54.82\% are predicted correctly whereas 45.18\% are not. Meanwhile, the 75 in the bottom left is the false negative (predict negative and it's false), and 1036 in the bottom right is the true negative (predict negative and it's true), which suggested that for all pairs of nodes with link existence (actual class = 1), 93.25\% are predicted correctly while 6.75\% are not.
We further calculate other evaluation metrics to quantify classification performance. The F1 score, which measures the test accuracy in an unbalanced class, was 0.74 for the predicted missing links (the range of the F1 score was [0, 1]), while the AUC was 0.84 for both training set and held-out test set. Basically, the AUC tells how capable the model is when distinguishing between classes. The higher the AUC, the better the model is. The problem of over-fitting is avoided because  the AUCs for both training set and test set are comparable.

\begin{table*}[t]
    \centering
    \caption{Confusion matrix in predicting 2014 and 2015 with 29 features. F1-score for 2014 is 0.65 and 0.65 for 2015. AUC for 2014 is 0.80 and 0.80 for 2015}    
    \begin{tabular}{@{}cccccc@{}}
    \multicolumn{2}{c}{}   &  \multicolumn{2}{c}{2014 test prediction on unseen network} &
    \multicolumn{2}{c}{2015 test prediction on unseen network}\\
    \noalign{\smallskip}\hline\noalign{\smallskip}
         \multirow{6}{*}{\begin{sideways} Actual Class \end{sideways}} &  &  0 &  1 & 0 & 1 \\
\hline\noalign{\smallskip}
         \rule{0pt}{23pt} & 0 & 42633 (TNR 61.73\%) & 26435 (FPR 38.27\%)  & 45735 (TNR 61.28\%) & 28893 (FPR 38.72\%)\\
         \rule{0pt}{23pt} & 1 & 1811 (FNR 15.17\%) & 10124 (TPR 84.83\%) & 2195 (FNR 16.43\%) & 11167 (TPR 83.57\%)\\
\noalign{\smallskip}\hline\noalign{\smallskip}
    \end{tabular}
    \label{tab:confusion_matrix__three years}
\end{table*}

\paragraph{Predicting entire network for 2014}
Once the trained model is converged, the learned parameters for the GNN model and the classification model can be used to predict the co-consideration network in the following years. As a test dataset, the car co-consideration network in 2014 is predicted. First, the 2014 car models set, which have an intersection with the 2013 car set and also have newly emerged cars, acts as the input of the prediction process without any link information. Then, through the adjacency prediction model, an approximate adjacency matrix based on the similarities of nodes is generated. Next, the node features and approximate prediction model are fed into the GNN model and followed by the classification model, the link existence of each pair of nodes is forecasted with a certain probability threshold. 

\begin{figure}[ht]
\centering
\includegraphics[width=0.45\textwidth]{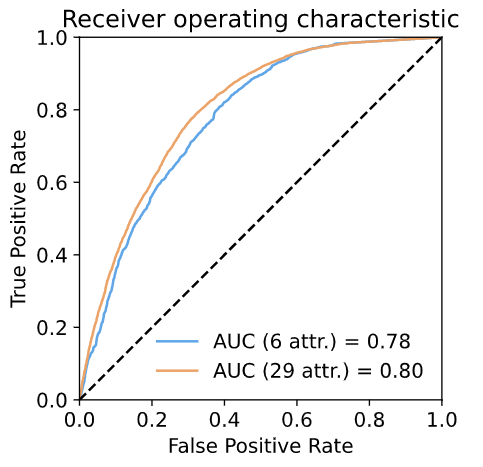}
\caption{AUC-ROC curve to predict 2014 co-consideration network with 6 attributes and 29 attributes}
\label{fig:auc_2014}
\end{figure}

Likewise, we computed the confusion matrix for the predicted 2014 co-consideration network in Table~\ref{tab:confusion_matrix__three years},and calculated the F1 score as 0.65. Furthermore, we scoped out the AUC-ROC curve (in Fig.~\ref{fig:auc_2014}) at various threshold settings. The overall AUC is 0.80.

\paragraph{Predicting entire network for 2015}
Hitherto, we have predicted the 2014 co-consideration network based on the training data in 2013. However, as 2014 succeeded 2013, the market structure did not change dramatically. Among 389 cars in 2013 and 403 cars in 2014, there are 296 cars in common. Therefore, to further assess the prediction capability for the model, we predict the 2015 co-consideration network using the trained model (2013 training data) with the car attributes and similarity-based adjacency matrix. 

The predicted results are recorded and evaluated in Table~\ref{tab:confusion_matrix__three years} and Fig.~\ref{fig:auc_2014}, where the F1 score is 0.65 and AUC is 0.80. Compared to the prediction results in 2014, the prediction in 2015 maintains an equivalent performance, which is an indication of model robustness. 

\paragraph{Predicting with six attributes}

In order to make a fair comparison of the previous traditional statistical network models (e.g., ERGMs), we used the same set of input attributes (only 6 attributes in~\cite{cui2020weighted}) and compared the AUC of each model. Besides, as previous studies used a subset of cars and did not make predictions for newly emerged car models, we took the intersection of 2013 and 2014 cars (296 cars in total) for our analysis. 

When only six car features were utilized in the training and prediction model, we obtained the prediction results for 2014 data for GNN and ERGM, respectively, in Table~\ref{tab:confusion_matrix_ergm}. In the confusion matrix, we observed that in ERGM, the true positive rate (the ratio of true positive to all actual positive) is 79.81\% and the true negative rate (the ratio of true negative to all actual negative) is 40.51\%. Both of the values are lower than those predicted by GNN. Furthermore, the F1 score of the ERGM is merely 0.31, which is almost half the F1 score of the GNN model. The AUC for ERGM prediction is 0.68, which is also less than the corresponding value of 0.78 for the GNN model. All of the evidence suggested that the prediction model of GNN performs better than the traditional statistical network models. 

\begin{table*}[!ht]
    \centering
    \caption{Confusion matrix in predicting 2014 with six features and 296 cars for using the GNN method and the ERGM method. F1 score is 0.60 for the GNN model and 0.31 for the ERGM model, and the AUC is 0.78 for the GNN model and 0.68 for the ERGM model.}    
    \begin{tabular}{@{}cccccc@{}}
    \multicolumn{2}{c}{}   &  \multicolumn{2}{c}{2014 prediction class GNN} &  \multicolumn{2}{c}{2014 prediction class ERGM}\\
    \noalign{\smallskip}\hline\noalign{\smallskip}
         \multirow{6}{*}{\begin{sideways} Actual Class \end{sideways}} &  &  0 &  1 & 0 & 1\\
\hline\noalign{\smallskip}
         \rule{0pt}{23pt} & 0 & 20336 (TNR 54.95\%) & 16675 (FPR 45.05\%) & 14993 (TNR 40.51\%) & 22018 (FPR 59.49\%)\\
         \rule{0pt}{23pt} & 1 & 867 (FNR 13.04\%) & 5782 (TPR 86.96\%) & 1384 (FNR 20.82\%) & 5265 (TPR 79.18\%)\\
\noalign{\smallskip}\hline\noalign{\smallskip}
    \end{tabular}
    \label{tab:confusion_matrix_ergm}
\end{table*}

Then we summarized all the AUCs to compare in Table ~\ref{tab:comparision_summary}. Notice that the ERGM with 29 attributes does not associate with an AUC value because the model does not get converged with so many attributes. Meanwhile, we did not run the six attributes prediction for the 2015 data on the GNN because the common car set for 2013 and 2014 is no longer suitable for the 2015 car market. It is apparent from the comparison that the GNN models perform better than the ERGM model with a higher AUC and F1 score, and GNN models can accommodate larger networks more design attributes and introduction of unseen nodes in the study of product relationship. 

\begin{table*}[ht]
    \centering
        \caption{Comparing train AUC and test AUC in different years, different models and different sets of attributes. AUC in link prediction. The goal is to predict the entire network (all existing and non-existing edges) in a 0/1 classification task}
    \begin{tabular}{ccccc}
    \toprule
      Number of attributes   &  Train AUC (2013) & Test AUC (2014) & Test AUC (2015) & Test AUC (ERGM)\\
      \midrule
       29 attributes & 0.84 & 0.80 & 0.80 & NA \\
        Six attributes & 0.81 & 0.78 & NA & 0.68 \\
        \bottomrule
    \end{tabular}
    \label{tab:comparision_summary}
\end{table*}

\begin{table*}[]
    \caption{Car attributes type and feature importance}
\pgfplotstablegetrowsof{\data}
\let\numberofrows=\pgfplotsretval
\renewcommand*{\arraystretch}{1}

\pgfplotstabletypeset[columns={Attribute,Variable Type,Importance,Sample Values},
  every head row/.style={before row=\toprule,after row=\midrule},
  every last row/.style={after row=[6ex]\bottomrule},
  columns/Attribute/.style={string type,column type=l,column name={\bf Attribute}},
  columns/Variable Type/.style={string type,column type=l,column name={\bf Variable Type}}, 
  columns/Importance/.style={
    column name={\bf Importance},
    assign cell content/.code={
   \ifnum\pgfplotstablerow=0
   \pgfkeyssetvalue{/pgfplots/table/@cell content}
   {\multirow{\numberofrows}{5cm}{\errplot}}%
   \else
   \pgfkeyssetvalue{/pgfplots/table/@cell content}{}%
   \fi
  }
 },
 columns/Sample Values/.style={string type,column type=l,column name={\bf Sample Values}},
]{\data}
    \label{tab:featureimportance}
\end{table*}

\subsection{Interpretability of Attributes}

To inspect the feature importance, we applied the permutation method to find the decrease in a model score when a single feature value is randomly shuffled. 
Permutation importance calculation repeats the process with multiple shuffles to ensure the accuracy. We ran 50 permutations for each feature in the training data and calculated the drop in performance. The results are shown in Table~~\ref{tab:featureimportance}. We found that the make of the car, the body type, and the segment are the most important attributes for the GNN to predict ties.

Table~\ref{tab:featureimportance} shows that 14 of the 29 attributes have no positive effect on the model prediction. Note that negative values are returned when a random permutation of a feature's values results in a better performance metric compared to the performance before a permutation is applied. This means the model does not rely on feature's which have negative values when predicting links for the training data. We observe that most continuous values, such as engine size, price, fuel consumption, and power do not have high importance. 

It is noteworthy that the permutation methods on feature importance can be applied to either training data or test data. In the end, one needs to decide whether one wants to know how much the model relies on each feature for making predictions (training data) or how much the feature contributes to the performance of the model on unseen data (test data).

\section{Implications for Design}
A car is an expensive commodity, and customers usually consider multiple options before deciding which car to buy. This decision may be influenced by many factors, such as the customer’s budget, driving needs, required and necessary features, the popularity of nearby car models, brand, past experience, the influence of cars owned or recommended by family and friends,~\etc

From a manufacturer's perspective, it is important to understand the market competition and develop strategies to improve their market share. The proposed model can support manufacturers in the following aspect:

First and foremost, the prediction capability of the GNN model facilitates the forecast of future market competition when a new car is introduced or the attributes of an existing car change. The model can be used by designers to anticipate the outcomes of a design change or a design release. For example, when a new car is released, the model can predict what other cars will be considered concurrently (co-consideration link existence). Therefore, designers or manufacturers can use this information to develop their design strategy. In addition, it is noticeable that the true positive rate for the prediction is over 80\% for all the results shown, which shows there is a considerably high probability that an actual link exists that will test positive. This indicates that  competition in the future can be well captured by the prediction model.

Secondly, the feature importance results shed light on understanding the key features in the co-consideration network formation. The results of the feature importance in Table~\ref{tab:featureimportance} show that some features, such as make, body type, import, lane assistance, third row, park assistance, and AWD, have a higher impact on the product co-consideration network, whereas other features, such as turbo and navigation, are not key factors in making predictions.
Knowing these factors and introducing interventions to change them for future product iterations can enable a car manufacturer to affect the competition relationships, leading to a larger market share. However, we should warn that it is imprudent to make definitive conclusions from regression models without real-world validation. Nevertheless, our analysis sheds light on key factors that customers may be considering while making their purchase decisions.

\section{Future Work and Limitations}
This work demonstrates the efficacy of GNNs in modeling products and the relationships between them. The findings of this study have several important implications for future practice, which will be discussed next.

\paragraph{Predict link strength in weighted and directed networks using GNN}
The current link existence prediction model consists of a GNN model and a classification model. Similarly, we can add a regression model as the downstream task instead of a classification model. The metrics of measuring the link strength prediction model could use root mean square error (RMSE) and mean absolute error (MAE). The prediction of link strength will enable designers to more precisely evaluate the effects of potential designs on  market demand compared to merely predicting the existence of links.

\paragraph{Predict network structure for multi-dimensional networks with heterogeneous links using GNN}
To further capture the relationship between customers and products, a multi-dimensional customer-product network can model the heterogeneous edges to model customers' considerations and choices simultaneously. In this work, we focused on undirected edges. However, future work will analyze directed edges to study the final choice of a customer from within a set of options.

\paragraph{Limitations}

First of all, this study is limited by the nature of survey data. The link existence between a pair of products is measured by the customers' consideration behavior. However, with the restriction of survey data, this study only samples a small portion of the real car market. The network we studied has a density of 14.73\%, which leads to an unbalanced dataset with most links classified as 0. To overcome the issue, we randomly selected a subset of samples from the original dataset to match the samples coming from both classes in the training process. 

Another limitation lies in the interpretability of feature importance. When two features are correlated and one of the features is permuted, the model will still have access to the feature through its correlated feature. This will result in a lower importance value for both features, whereas they might actually be important. The problem is common in many interpretable machine learning problems, and our work is no exception to it.

Thirdly, a notable drawback of GraphSAGE is that sampled nodes might appear multiple times, thus potentially introducing a lot of redundant computation. With the increase of the batch size and the number of samples, the number of redundant computations increases as well. Moreover, despite having nodes in memory for each batch, the loss is computed on a single batch of them, and, therefore, the computation for the other nodes is also in some sense wasted. Further, the neighborhood sampling used in GraphSAGE is effective in improving computing and memory efficiency when inferring a batch of target nodes with diverse degrees in parallel. Despite this advantage, the default uniform sampling can suffer from high variance in training and inference, leading to sub-optimum accuracy. While new architectures inspired by GraphSAGE attempt to reduce computation time and performance variation, we did not focus on finding the best architecture for improving the computational efficiency, as it was not central to our focus area.

\section{Conclusions}

We present a systematic method to study and predict relationship between products by using the inductive graph neural network models. 

This paper makes the following key contributions:
\begin{enumerate}
    \item We show that neural network models, which can embed each node of a graph into a real vector, can capture node feature and graph structure information, to enable machine learning applications on complex networks. This is the first attempt of implementing GNNs to predicting product relationships.
    \item We show that GNN models have better link prediction performance than ERGMs, both for held-out links from the same year and predicting the entire network structure for future years.
    \item We overcome a limitation of GNN by proposing a new method to predict links between unseen cars for future years.
    \item We show the scalability of the GNN method by modeling the effect of a large number of continuous and categorical attributes on link prediction.
    \item We use permutation-based methods to find the importance of attributes to help design decisions.
\end{enumerate}

In future work, we aim to make predictions on the product relationship strength and extend the current work on more complex network structures to investigate the relationship among customers and products.

\bibliographystyle{asmems4}

\bibliography{asme2e}

\end{document}